\documentclass[11pt,draftcls,letterpaper,onecolumn]{IEEEtran}

\usepackage[dvips]{color}
\usepackage{epsf}
\usepackage{times}
\usepackage{epsfig}
\usepackage{graphicx}
\usepackage{url}
\usepackage{amsmath}
\usepackage{amssymb}
\usepackage{amsxtra}
\usepackage{algorithm}
\usepackage{algorithmic}
\usepackage{cite}
\usepackage{listings}
\usepackage{verbatim}
\usepackage{pseudocode}
\usepackage{bm}
\usepackage{amsthm}

\theoremstyle{plain}

\newtheorem{theorem}{Theorem}

\newtheorem{lemma}{Lemma}

\theoremstyle{definition}

\theoremstyle{remark}
\newtheorem*{remark}{Remark}

\begin{document}

\title{Nearly-tight Bounds on Linear Regions of Piecewise Linear Neural Networks}

\author{Qiang~Hu, Hao Zhang% <-this % stops a space
\thanks{Qiang Hu and Hao Zhang were with the Department
of Electrical  Engineering, Tsinghua University, Beijing, China, 
e-mail: huq16@mails.tsinghua.edu.cn.}}

\maketitle

\begin{abstract}
 The developments of deep neural networks (DNNs) in recent years have ushered a brand new era of artificial intelligence. DNNs are proved to be excellent in solving very complex problems, e.g., visual recognition and text understanding, to the extent that their performance competes with or even surpasses humans. Despite inspiring and encouraging success of DNNs, thorough theoretical analyses still lack to unravel mysteries of their magics. The design of DNN structures is dominated by empirical results in terms of network depth, number of neurons and activation functions. A few of remarkable works published recently in an attempt to interpret DNNs have established the first glimpses of their internal mechanisms. Nevertheless, research on exploring how DNNs operate is still at the initial stage with plenty of room for refinement. In this paper, we extend precedent research on linear region bounds of neural networks with piecewise linear activations (PLNNs). We present (i) the exact maximal number of linear regions computable by single layer PLNNs; (ii) an upper bound on the number of linear regions for multi-layer PLNNs; and (iii) tighter upper and lower bounds on number of the linear regions of rectifier networks. The derived bounds indicate that deep models are more expressive than shallow counterparts, and non-linearity of activation functions also has a great impact on expressiveness of neural networks.
\end{abstract}

\begin{IEEEkeywords}
Deep neural network, input space partition, piecewise linear function, learning theory.
\end{IEEEkeywords}

% \fi
\IEEEpeerreviewmaketitle

\section{Introduction}
The emergence of deep neural networks (DNNs) has greatly promoted the development of artificial intelligence due to their state-of-the-art results in computer vision, speech recognition and a variety of other machine tasks \cite{Krizhevsky2012,Hinton2012,Goodfellow2014}. Some popular networks proposed in recent years such as GoogleNet \cite{Szegedy2015} and ResNet \cite{He2016} have achieved record breaking accuracies on image classification. Despite unprecedented success, the design of DNNs mainly relies on empirical results without a solid theoretic basis. It's a common view that such powerful capabilities of DNNs lie in great depth of layers and the use of non-linear activation functions, while the underlying reasons are still not fully investigated. Fortunately, a few of researchers have made great contributions to unveil the internal mechanisms of DNNs in a theoretical perspective \cite{Cybenko1989,Anthony1999,Pascanu2013,Montufar2014,Bianchini2014,Raghu2017,Arora2018,Serra2018}. 

A chief concern about theoretical analysis is mathematically quantifying the expressiveness of DNNs. It has been shown that DNNs are exponentially efficient than shallow counterparts at modeling certain families of continuous functions \cite{Delalleau2011}. The compositional property of DNNs enables higher layers reuse ingredients provided by lower layers to build gradually more complex functions, while shallow models can only construct the target detectors based on the primitives learned by a single layer. As a result, the depths of modern neural network architectures on vision tasks always go beyond one hundred to extract complicated features from images. 

Non-linearity of activation functions also has a great influence on the modeling capabilities. Nowadays, piecewise linear (PWL) functions, such as the rectifier activation, have been mostly common choices in the design of deep models. Most of current marvelous and impressive structures of DNNs involve piecewise linear activations. Glorot et al. have proved that rectifier activations can reduce the complexity of optimization problem compared with traditional bounded smooth activations, such as sigmoid and tanh activations \cite{Glorot2011}. The success of PLNNs has driven us to figure out the theoretical basis of strong expressiveness of DNNs. 

 The expressiveness of DNNs with rectifier activations has been intensively studied and a series of theoretical results have been proposed \cite{Pascanu2013, Montufar2014, Arora2018,Serra2018}. Such analyses are based on the fact that PLNNs are indeed a set of PWL functions given that a composition of PWL functions is still a PWL function. The input space is divided into several linear regions by these PWL functions and each region corresponds to a specific linear function. The more linear regions are, the more complex functions PLNNs can model. The expressiveness of PLNNs can be quantified by counting the number of linear regions. Pacanu et al. have shown that the number of linear regions partitioned by deep rectifier networks is exponentially more than that of shallow counterparts with the same number of hidden neurons in asymptotic limit of layers. Such results are significantly improved by Mont\'ufar et al.. The corresponding upper and lower bounds on the maximal number of linear regions computable by rectifier networks and maxout networks \cite{Goodfellow2013} are obtained. Raghu et al. further improved the upper bound on the number of linear regions of rectifier networks, and this upper bound is asymptotically close to lower bound of Mont\'ufar el al. on certain conditions. Subsequently, Arora et al. improved the lower bound and provided a family of rectifier networks that achieve an exponential number of regions for fixed size and depth. Finally, Serra et al. further tightened both of upper and lower bounds on the number of linear regions for rectifier networks.

This paper extends and improves the results mentioned above by deriving the bounds on the number of linear regions computable by PLNNs. Our main contributions are summarized in the following lists:

\begin{itemize}
\item We extend the analysis of rectifier networks to PLNNs and provide the exact maximal number of linear regions as well as corresponding asymptotic expansions computed by the PWL functions of single layer PLNNs. This bound grows asymptotically polynomially in the number of hidden neurons and the number of pieces of linear activations when input dimension is constant.

\item We derive an upper bounds on the maximal number of linear regions for deep PLNNs. This bound is exponentially increased with the depth of PLNNs compared with shallow counterparts when the number of hidden neurons is fixed. It provides an insight on how the depth of PLNNs and the non-linearity of activation functions effect expressiveness of deep PLNNs.

\item We tighten upper and lower bounds on the maximal number of linear regions for rectifier networks by considering constraints from geometric properties of linear regions computed by PWL functions and activation patterns when some neurons are nonactive. Additionally, this upper and lower bounds have significantly improved compared with previous bounds.  
\end{itemize}

\section{Preliminaries}
In this paper, fully connected PLNNs $\mathcal{N}$ with $n_{0}$ input neurons, $L$ hidden layers, and $m$ output neurons are considered. Denote by $\mathbf{x}=[x_{1},x_{2},\ldots,x_{n_{0}}]^{T}$ the input vector to $\mathcal{N}$, and $\mathbf{x}\in\mathcal{X}$, where $\mathcal{X}\subseteq\mathbb{R}^{n_{0}}$ is an nonempty subset of $n_{0}$-dimension input space. Assume that each hidden layer $l\in[L]$ has $n_{l}$ neurons and the number of output units is $n_{L+1}$. The computation of $\mathcal{N}$ proceeds in a feed-forward way in the form of a piecewise linear (PWL) function $\mathcal{F}_{\mathcal{N}}:\mathbb{R}^{n_{0}}\rightarrow\mathbb{R}^{n_{L+1}}$ given by
\begin{align}
\mathcal{F}_{\mathcal{N}}(\mathbf{x},\theta)=f_{L+1}\circ h_{L}\circ f_{L}\circ\cdots\circ h_{1}\circ f_{1}(\mathbf{x}),
\end{align}
where $f_{l}$ is a linear function and $h_{l}$ is a PWL activation function. Let $\mathbf{W}^{l}\in\mathbb{R}^{n_{l}\times n_{l-1}}$ represents the weight matrix and $\mathbf{b}^{l}\in\mathbb{R}^{n_{l}}$ be the bias vector assigned to hidden layer $l$. All the $\mathbf{W}^{l}$ and $\mathbf{b}^{l}$ compose the parameter $\theta$ for $l\in[L]$. Let $\mathbf{x}^{l}=[x_{1}^{l},\ldots,x_{n_{l}}^{l}]^{T}$ be the output of the $l$-th layer. Given the output from previous layer, the pre-activation of the $l$-th layer is given by
\begin{align}
\mathbf{z}^{1}=f_{1}(\mathbf{x})&=\mathbf{W}^{1}\mathbf{x}+\mathbf{b}^{1},\\
\mathbf{z}^{l}=f_{l}(\mathbf{x}^{l-1})&=\mathbf{W}^{l}\mathbf{x}^{l-1}+\mathbf{b}^{l}.
\end{align}
where $\mathbf{z}^{l}=[z_{1}^{l},\ldots,z_{n_{1}}^{l}]^{T}$ and $z_{i}^{l}$ is weighted sum of real-valued activations $x_{1}^{l-1},\ldots,x_{n_{l}}^{l-1}$ of layer $l-1$ for $i\in[n_{l}]$ and $l>1$. Additionally, $\mathbf{z}^{1}$ is a linear combination of input values $x_{1},\ldots,x_{n_{0}}$. Applying PWL activation function $h_{l}$ to $\mathbf{z}^{l}$, the output activation vector of layer $l$ is given by
\begin{align}
\mathbf{x}^{l}=h_{l}(\mathbf{z}^{l})=[h_{l}(z_{1}^{l}),\ldots,h_{l}(z_{n_{l}}^{l})]^{T}.
\end{align}
For the neuron $i\in[n_{l}]$ of layer $l$, the explicit form of PWL activation function $h_{l}(z_{i}^{l})$ is written as 
\begin{align}\label{activation}
h_{l}(z_{i}^{l})=\left\{\begin{array}{cc}
r_{1}z_{i}^{l}+t_{1}, & \mathrm{if}\ z_{i}^{l}\in A_{1}\\
r_{2}z_{i}^{l}+t_{2}, & \mathrm{if}\ z_{i}^{l}\in A_{2}\\
\vdots & \vdots\\
r_{p}z_{i}^{l}+t_{p}, & \mathrm{if}\ z_{i}^{l}\in A_{p+1}
\end{array}\right.
\end{align}
where $p\geq 1$ is a constant integer representing the number of breakpoints, $\{A_{1},\ldots,A_{p+1}\}$ are corresponding disjoint intervals satisfying $\bigcup_{i=1}^{p+1}A_{i}=\mathbb{{R}}$, $\{r_{1},\ldots,r_{p+1}\}$ are constant slopes, and $\{t_{1},\ldots,t_{p+1}\}$ are constant intercepts. Let $\{e_{1},\ldots,e_{p}\}$ be a set of disjoint breakpoints that partition $\mathbb{R}$. It follows that $A_{1}=(-\infty,e_{1}],A_{2}=(e_{1},e_{2}],\ldots,A_{p+1}=(e_{p},\infty)$. Generally, rectified linear unit (ReLU) is the most commonly used form of hidden neurons in PLNNs. Its activation function has only two pieces $A_{0}=(-\infty, 0], A_{1}=(0,\infty)$ with $h_{l}(z_{i}^{l})=0$ and $h_{l}(z_{i}^{l})=z_{i}^{l}$, respectively. The simplified form of ReLU activation function is defined as $f_{relu}(x)=\max\{0,x\}$.

As mentioned by Mont\'ufar et al. \cite{Montufar2014}, the number of computable PWL functions from $\mathcal{F}_{\mathcal{N}}$ is dependent on the structure of $\mathcal{N}$, i.e., neuron arrangement $\{n_{0},n_{1},\ldots,n_{L}\}$ and depth $L$. Moreover, the number of functions computable from $\mathcal{F}_{\mathcal{N}}$ is equivalent to count their numbers of linear regions of input space partitioned by these PWL functions. Given a PWL function $f:\mathbb{R}^{n_{0}}\rightarrow\mathbb{R}^{n_{L}}$, a linear region is a maximal connected open subset of the input space $\mathbb{R}^{n_{0}}$. In the next sections, we will compute the number of linear regions for PLNNs with one single hidden layer (shallow PLNNs) and upper bound for PLNNs with multiple hidden layers (deep PLNNs). 

In the remainder of this section, we introduce the definition of activation pattern of PLNNs. 
Given a fixed PLNN $\mathcal{N}$, the pre-activation $z_{i}^{l}$ of the $i$-th neuron of layer $l$ is computed from the input vector $\mathbf{x}$ for $i\in[n_{l}]$. $z_{i}^{l}$ further determines the form of PWL activation function $h_{l}(z_{i}^{l})$, namely, the linear function in Eq.\eqref{activation} to apply. The status of each neurons is encoded into $p$ different states based on the number of linear function available in Eq.\eqref{activation}. Denote by $s_{i}^{l}\in\{1,\ldots,p+1\}$ the activation pattern of each neuron of layer $l$ such that $s_{i}^{l}=q(q\in\{1,\ldots,p+1\})$ if and only if $z_{i}^{l}\in A_{q}$. The activation pattern of all the neurons of layer $l$ is then given by a vector $\mathbf{s}^{l}=[s_{1}^{l},\ldots,s_{n_{l}}^{l}]^{T}$. The activation pattern of neurons up to layer $l\leq L$ is an aggregate vector $\mathbf{s}_{l}=[(\mathbf{s}^{1})^{T},\ldots,(\mathbf{s}^{l})^{T}]^{T}$. Let $\mathcal{S}^{l}\subseteq\{1,\ldots,p+1\}^{\sum_{i=1}^{l}n_{i}}$ be the activation set that $\mathbf{s}_{l}$ belongs to, and $\mathcal{S}^{L}$ specifies all the possible activation patterns of $\mathcal{N}$. The inputs located at the same linear region partitioned by the PWL functions from $\mathcal{F}_{\mathcal{N}}$ correspond to the same activation pattern of $\mathcal{S}^{L}$.

\section{Shallow PLNNs}
This section mainly analyzes the number of linear regions computed by the PWL functions of shallow $\mathcal{N}$ with only one hidden layer composing of $n_{0}$ inputs and $n_{1}$ hidden neurons. For every neuron $i\in[n_{1}]$ in hidden layer, $z_{i}^{1}$ decides which one of $p+1$ linear functions in Eq.\eqref{activation} is activated. The boundaries between these $p+1$ activation patterns are given by $n_{1}$ groups of $p$ parallel hyperplanes $\{H_{i,j}\}_{i\in[n_{1}],j\in[p]}$ in $n_{0}$-dimension input space. Each $H_{i,j}$ is defined by $(\mathbf{w}_{i}^{1})^{T}\mathbf{x}+\mathbf{b}_{i}^{1}-e_{j}=0$, where $\mathbf{w}_{i}^{l}$ is the $i$-th row of $\mathbf{W}^{1}$ and $b_{i}^{1}$ is the $i$-th entry of $\mathbf{b}^{1}$, respectively. These parallel hyperplanes separate the input space into disjoint linear regions with each region corresponding to a distinct activation pattern. The linear function activated in Eq.\eqref{activation} is dependent on the linear region $z_{i}^{1}$ located at. 

It follows that these hyperplanes form a hyperplane arrangement $\mathcal{A}=\{H_{i,j}\}_{i\in[n_{1}],j\in[p]}$. The number of activation patterns of shallow PLNNs is equal to the number of linear regions partitioned by the hyperplanes from $\mathcal{A}$. Formally, a linear region of $\mathcal{A}$ is a connected component formed by the complement $\mathbb{R}^{n_{0}}\backslash\left(\cup_{i\in[n_{1}],j\in[p]}H_{i,j}\right)$. According to Zaslavsky's theorem \cite{Zaslavsky1975}, the maximal number of linear regions generated by an arrangement of $n_{1}$ hyperplanes in $\mathbb{R}^{n_{0}}$ is given by $\sum_{n=0}^{n_{0}}\binom{n_{1}}{n}$. Furthermore, this maximal number of linear regions holds if and only if all these hyperplanes are in general positions. Unfortunately, Zaslavsky's theorem can not be directly applied to shallow PLNNs due to parallelisms of some hyperplanes in $\mathcal{A}$. It's only valid in the special case when $p=1$, e.g., rectifier networks with a single hidden layer \cite{Pascanu2013}. 

As the number of output neurons has no influence on the maximal number of linear regions (See \cite{Pascanu2013}, Lemma2), the maximal number of linear regions partitioned by the hyperplanes of $\mathcal{A}$ is uniquely determined by $n_{1}$ and $p$. The following theorem derives the maximal number of linear regions of shallow PLNNs. 

\begin{theorem}\label{theorem1}
Define $\mathcal{R}\left(\mathcal{A}\right)$ as the maximal number of linear regions generated by $\mathcal{A}$, and it is given by  
\begin{align}\label{th1}
\mathcal{R}\left(\mathcal{A}\right)=\sum_{i=0}^{n_{1}}\binom{n_{1}}{i}\left(p-1\right)^{i}\sum_{n=0}^{n_{0}-i}\binom{n_{1}-i}{n}.
\end{align}
When $n_{1}\leq n_{0}$, $\mathcal{R}\left(\mathcal{A}\right)=\left(p+1\right)^{n_{1}}$. When $n_{1}>n_{0}$ and $n_{0}=O(1)$, $\mathcal{R}\left(\mathcal{A}\right)$ behaves as $\Theta\left((p+n_{1})^{n_{0}}\right)$ asymptotically.
\end{theorem}

\begin{IEEEproof}
See Appendix \ref{A}. 
\end{IEEEproof}

\begin{remark}
It's easy to see that the maximal number of activation patterns of shallow PLNNs is equal to $\mathcal{R}\left(\mathcal{A}\right)$. Generally, such result can extend to the maximal number of linear regions partitioned by any $n$ groups of $p$ parallel hyperplanes in $d$-dimension space. Denote by $\mathcal{R}\left(d, n\right)$ be the corresponding number of linear regions, and $\mathcal{R}\left(d,n\right)$ is equal to the right-hand side (RHS) of Eq.\eqref{th1} with $n_{0}$ and $n_{1}$ replaced with $d$ and $n$.

Intuitively, the maximal number of linear regions partitioned by the PWL functions computed by shallow PLNNs is upper bounded by $(p+1)^{n_{1}}$. The derived bound in Theorem \ref{theorem1} indicates that the number of linear regions of shallow PLNNs actually grows polynomially in $n_{1}$ and $p$ if $n_{0}$ is constant and $n_{1}>n_{0}$, instead of growing exponentially in $n_{1}$ and $p$. The reason of polynomial constraint results from linear dependence of normal vectors of the hyperplanes from $\mathcal{A}$ when the number of hyperplanes is large than the dimension of input space (See \cite{Anthony1999}, Theorem 3.4). Such linear dependence constrains the degree of freedom of parameter space of PLNNs. In this case, exponential upper bound on the number of linear regions isn't able to achieve. Besides, the parallelism of some hyperplanes in $\mathcal{A}$ also lowers the number of possible linear regions. As a result, the number of linear regions computed by shallow PLNNs is polynomially upper bounded on $p$ and $n_{1}$ with the degrees no larger than $n_{0}$.
\end{remark}

The result in Theorem \ref{theorem1} suggests that non-linearity of activation functions has a great effect in the expressiveness of neural network, which is not observed in the case of rectifier networks. The neural networks with more complex activation functions are considered to be more expressive. In the next, the number of linear regions computed by the PWL functions of deep PLNNs is discussed.

\section{Deep PLNNs}
This section mainly focus on quantifying the expressiveness of multiple layer $\mathcal{N}$ $(L\geq 2)$ in terms of the number of linear regions. We derive an upper bound on the number of linear regions as well as its corresponding asymptotic expression.  
\subsection{Problem Definitions}
Denote by $r_{i}^{l}$ and $t_{i}^{l}$ the slope and the intercept of activated linear function for every neuron of layer $l$ for $i\in[n_{l}]$ and $l\in[L]$. It follows that $r_{i}^{l}=r_{q}$ and $t_{i}^{l}=t_{q}$ if and only if $s_{i}^{l}=q\ (q\in\{1,\ldots,p+1\})$. In this way, the activation patterns lie in the slopes and the intercepts of activated linear functions of hidden neurons. Aggregate all the activated slopes and intercepts of layer $l$ into $\mathbf{r}^{l}=[r_{1}^{l},\ldots,r_{n_{l}}^{l}]^{T}$ and $\mathbf{t}^{l}=[t_{1},\ldots,t_{n_{l}}^{l}]^{T}$, respectively. The activation function of layer $l$ can be rewritten as
\begin{align}
h_{l}(\mathbf{z}^{l})=\mathrm{diag}(\mathbf{r}^{l})\mathbf{z}^{l}+\mathbf{t}^{l}.
\end{align}
Hence, the pre-activation $\mathbf{z}^{l+1}$ of next hidden layer $l+1$ is given by
\begin{align}
\mathbf{z}^{l+1}=\mathbf{W}^{l+1}\mathbf{x}^{l}+\mathbf{b}^{l+1}=\mathbf{W}^{l+1}\mathrm{diag}(\mathbf{r}^{l})\mathbf{z}^{l}+\mathbf{W}^{l+1}\mathbf{t}^{l}+\mathbf{b}^{l+1}=\tilde{\mathbf{W}}^{l+1}\mathbf{z}^{l}+\tilde{\mathbf{b}}^{l+1}
\end{align}
where $\tilde{\mathbf{W}}^{l+1}=\mathbf{W}^{l+1}\mathrm{diag}\left(\mathbf{r}^{l}\right)$ and $\tilde{\mathbf{b}}^{l}=\mathbf{W}^{l+1}\mathbf{t}^{l}+\mathbf{b}^{l+1}$. By expanding $\mathbf{x}^{l}$, $\mathbf{z}^{l+1}$ is recursively rewritten as
\begin{align}\label{eq7}
\mathbf{z}^{l+1}&=\prod_{i=1}^{l+1}\tilde{\mathbf{W}}^{i}\mathbf{x}+\sum_{i=1}^{l}\left(\prod_{j=1}^{i}\tilde{\mathbf{W}}^{l+1-j}\right)\tilde{\mathbf{b}}^{l+1-j}+\tilde{\mathbf{b}}^{l+1}\nonumber\\
&=\hat{\mathbf{W}}^{l+1}\mathbf{x}+\hat{\mathbf{b}}^{l+1}
\end{align}
where $\tilde{\mathbf{W}}^{1}=\mathbf{W}^{1}$, $\tilde{\mathbf{b}}^{1}=\mathbf{b}^{1}$, and $\hat{\mathbf{W}}^{l+1}=\prod_{i=1}^{l+1}\tilde{\mathbf{W}}^{i}$ and $\hat{\mathbf{b}}^{l+1}=\sum_{i=1}^{l}\left(\prod_{j=1}^{i}\tilde{\mathbf{W}}^{l+1-j}\right)\tilde{\mathbf{b}}^{l+1-j}+\tilde{\mathbf{b}}^{l+1}$ are the equivalent coefficient matrix and the bias with respect to $\mathbf{x}$. The explicit form of $\mathcal{F}_{\mathcal{N}}(\mathbf{x})$ is then given by
\begin{align}\label{eq8}
\mathcal{F}_{\mathcal{N}}(\mathbf{x})=\hat{\mathbf{W}}^{L+1}\mathbf{x}+\hat{\mathbf{b}}^{L+1}.
\end{align}

The result indicates that $\mathcal{F}_{\mathcal{N}}(\mathbf{x})$ is a linear classifier dependent on $\mathbf{x}$. As $\mathbf{x}$ ranges over the input space, each distinct activation pattern corresponds to a distinguish linear classifier. More importantly, despite infinite number of instances $\mathbf{x}$, the number of equivalent classifiers is subject to the number of activation patterns of $\mathcal{N}$. Generally, $\left(p+1\right)^{N}$ is a loosely upper bound on the maximal number of linear regions of $\mathcal{N}$, where $N=\sum_{l=1}^{L}n_{l}$. To gain a better understanding of expressiveness of $\mathcal{N}$, a more precisely tight upper bound is derived in the next. In particular, rectifier networks need to be treated differently due to zero-output properties of hidden neurons during the feedforward transmission process.

\subsection{Number of Linear Regions}\label{sec1}
Counting the number of linear regions is more complicated in the case of deep PLNNs. The partition of input space is a recursive process layer by layer. At each hidden layer, newly generated linear regions are obtained from partitioning the linear regions computed by the PWL functions of previous layer. Let $\mathcal{R}^{l}$ be the set containing all the linear regions up to layer $l$ and $\mathcal{R}^{0}$ is $n_{0}$-dimension input space. Every linear region of $\mathcal{R}^{l}$ corresponds to a specific activation pattern. Moreover, the linear regions of $\mathcal{R}^{l+1}$ are obtained by recursively partitioning the linear regions of $\mathcal{R}^{l}$. Given a linear region $R\in\mathcal{R}^{l}$ corresponding to an activation vector $\mathbf{s}_{l}\in\mathcal{S}^{l}$, it would be further partitioned by a set of hyperplanes $P_{\mathbf{s}_{l}}=\{(\hat{\mathbf{w}}_{i}^{l+1})^{T}\mathbf{x}+\hat{b}_{i}^{l+1}-e_{j}=0\}_{i,j}$ for $i\in[n_{l+1}]$ and $j\in[p]$. The normal vector $\hat{\mathbf{w}}_{i}^{l+1}$ and the bias $\hat{b}_{i}^{l+1}$ are varied with the activation vector $\mathbf{s}_{l}$. Generally, only a subset of these hyperplanes that intersect with interior of $R$ is effective. Denote by $\mathcal{N}_{R}^{l}$ the maximal number of linear regions and these linear regions are partitioned by a set of the hyperplane set $\{P_{\mathbf{s}_{l}}\}_{\mathbf{s}_{l}\in\mathcal{S}^{l}}$. The number of linear regions up to layer $l$ is given by
\begin{align}\label{recur}
\mathcal{N}^{l}=\sum_{R\in \mathcal{R}^{l-1}}\mathcal{N}_{R}^{l-1},\quad\mathcal{N}_{R}^{0}=1,\;\mathrm{for\;each\;region}\;R\subseteq\mathbb{R}^{n_{0}}.
\end{align}  
 
The recursion formula Eq.\eqref{recur} counts the number of linear regions by moving along the branches of a tree rooted at $\mathcal{R}^{0}$. As layer of $\mathcal{N}$ deepens, the height of tree increases and the number of linear regions grows exponentially with it. Based on the recursive relationship, the maximal number of linear regions computed by $\mathcal{N}$ is $\mathcal{N}^{L}=\sum_{R\in R^{L-1}}\mathcal{N}_{R}^{L-1}$. In the next, we will give detailed analysis of upper bound on the number of linear regions of deep PLNNs. 

\subsection{Upper Bound for Deep PLNNs}

The upper bound on the number of linear regions of deep PLNNs is derived in the following theorem. 

\begin{theorem}\label{theorem2}
Consider an $L(L\geq 2)$-layer neural network $\mathcal{N}$ with $p+1$ piecewise linear activations, $n_{l}$ hidden neurons at each layer $l$, and $n_{0}$-dimension input. The maximal number of linear regions computed by $\mathcal{N}$ is upper bounded by
\begin{align}
\mathcal{N}^{L}\leq\prod_{l=1}^{L}\mathcal{R}\left(d_{l},n_{l}\right)
\end{align}
where $d_{l}=\min\{n_{0},n_{1},\ldots,n_{l}\}$.
\end{theorem} 
\begin{IEEEproof}
See Appendix \ref{B}.
\end{IEEEproof} 
\begin{remark}
See Appendix \ref{C}.
\end{remark}

This upper bound is a polynomial with its degree no larger than the product of the input dimension and the depth of networks. Such upper bound is in accordance with the conclusion of Theorem \ref{theorem1} when $L=1$ and exponentially growing with $L$. Moreover, this bound is tight when $n_{0}=1$ and $p=1$ if $n_{l}\geq 3$ for $l\in[L]$ (See \cite{Serra2018}, Theorem 7). 

\subsection{Bounds for Rectifier Networks}
Rectifier networks is a special form of PLNNs. Single layer rectifier networks make no differences with shallow PLNNs in terms of the number of linear regions by setting $p=1$. As for multiple layer rectifier networks, counting the number of linear regions they generated should be treated differently from any other PLNNs. Serra et al. have proposed upper and lower bounds on the number of linear regions computed by the PWL functions of rectifier networks \cite{Serra2018}. We tighten such bounds by considering the properties of activation patterns of rectifier units at a more detailed level. 

Generally, the linear regions partitioned by the PWL functions of deep PLNNs are classified into two types, bounded polyhedron and unbounded polyhedron. We refer bounded polyhedron as polytope and unbounded polyhedron simply as polyhedron. It follows that the activation patterns of polyhedrons are complementary as they stretch to infinity and every polyhedron has a counterpart in opposite axial directions. In order to count the number of linear regions of rectifier networks, we need be aware that how ploytopes and polyhedrons are distributed in the input space. 
\begin{lemma}\label{lemma1}
Consider $m$ hyperplanes in $\mathbb{R}^{n_{0}}$ defined by  $\{\mathbf{w}_{i}^{T}\mathbf{x}+b_{i}=0\}$, where $\mathbf{w}_{i}\in\mathbb{R}^{d}$ is the normal vector, $\mathbf{x}$ is $d$-dimension real-valued input, and $b_{i}\in\mathbb{R}$ is the bias for $i\in[m]$. Let $\mathbf{W}=[\mathbf{w}_{1},\ldots,\mathbf{w}_{m}]$ be the matrix containing all the normal vectors and $r$ be the rank of $\mathbf{W}$. For $m,r\geq 1$, the maximal number of polyhedrons partitioned by these hyperplanes is $2\sum_{n=0}^{r-1}\binom{m-1}{n}$. Each polyhedron has a counterpart satisfying their corresponding active neurons added up to $m$.
\end{lemma}
\begin{IEEEproof}
See Appendix \ref{D}.
\end{IEEEproof} 
If $m\leq d$, the maximal number of polyhedrons is $2^{m}$, which is directly equal to the maximal number of linear regions. In this case, all the hyperplanes intersect at one point and all the linear regions turn into cones if the intersected point is original. If $m>r$, some linear regions must be polytopes if the input space is maximally partitioned, and the maximal number of polytopes is simply given by $\binom{m-1}{r}$. Beyond that, the numbers of active neurons of polyhedrons are complementary. Such property constrains the number of linear regions generated by rectifier networks. For example, if there is a polyhedron that all the neurons are active, we must have another complementary polyhedron that all the neurons are non-active. 

The upper bound on the number of linear regions of deep PLNNs is obtained by recursively bounding the number of subregions within a linear region as layer increased. The biggest difference between deep PLNNs and rectifier networks is that the number of active hidden neurons of rectifier networks would greatly affect the dimension of output space of hidden layers, which is also the input space of next hidden layers. The information flow through hidden neurons is cut off once these neurons are nonactive and output zero. As a result, the dimension of output space is lower than that of input space if the number of active neurons is insufficient for the hidden layer. The achievable number of linear regions computed by the PWL functions of subsequent layers is decreased with the dimension of output space. Hence, in order to upper bound the number of linear regions, we need to activate as many hidden neurons as possible. Meanwhile, Lemma \ref{lemma1} implies that the polyhedrons are complementary in terms of the number of active neurons. We should consider such constrain when counting the number of linear regions of rectifier networks. 

The lower bound on the number of linear regions of deep PLNNs is improved as well. According to Mont\'{u}far et al.'s theorem (See \cite{Montufar2014}, Theorem 4), we can construct fictitious intermediary layers connecting two hidden layers. The units of intermediary layer receive the outputs from previous hidden layer and compute the inputs passed to the next hidden layer. To make all the intermediary units independent through the whole rectifier networks, the minimal value of the number of hidden neurons up to layer $L-1$ is set as the initial dimension of the first intermediary layer, i.e., input layer, which is denoted by $\tilde{d}=\min\{n_{0},n_{1},\ldots,n_{L-1}\}$. Therefore, all the intermediary layers are guaranteed to have $\tilde{d}$ units. Denote by $d_{l}\leq\tilde{d}$ the dimension of intermediary layer for $l\in[L]$. It's the number of effective intermediary units of layer $l$ as some intermediary units may be zero if all the connected hidden neurons output zero. It follows that the initial dimension satisfies $d_{1}=\tilde{d}$, i.e. the dimension of input layer is $\tilde{d}$. Assume that the first $\tilde{d}$ neurons of input layer are selected as the inputs.  

Similar to the processing of Mont\'{u}far et al., the neurons of hidden layer $l$ are partitioned into $\tilde{d}$ different groups with cardinality equal to $\lfloor n_{l}/\tilde{d}\rfloor$ for $1\leq l\leq L-1$. The neurons of the last hidden layer aren't considered as they have no subsequent hidden layers to act on. The number of remaining neurons of layer $l$ is denoted by $r_{rem}^{l}=n_{l}-\lfloor n_{l}/\tilde{d}\rfloor\tilde{d}$. These neurons are assigned to a subset of $\tilde{d}$ groups one by one starting from the first group. Consequently, the first $r_{rem}^{l}$ groups contain $\lfloor n_{l}/\tilde{d}\rfloor+1$ neurons and the rest have $\lfloor n_{l}/\tilde{d}\rfloor$ neurons. Each group connects to a separate intermediary unit and are mutually independent. The activation patterns of neurons of each group are only determined by the connected intermediary unit. We can construct a zigzag pattern from $[0,1]$ to $[1,0]$ within each group to map all the inputs from the intermediary units belong to different activation patterns to single output range $[0,1]$. All the activations from different groups constitute a $d_{l}$-dimension cube. Such cube become the outputs of next intermediary layer, which is equivalent to a range finite $d_{l}$-dimension space. In this way, all the partitions of next hidden layer operated on this cube are replicated within all the linear regions of previous hidden layer. The number of activation patterns of each group is equal to the number of slopes of the zigzag pattern. Bear these in mind, we recursively bound the number of linear regions of rectifier networks, which is given by the following theorem. 
\begin{theorem}\label{theorem3}
Consider an $L(L\geq 2)$-layer rectifier network with $n_{0}$-dimension input and $n_{l}$ hidden neurons of layer $l$. Denote by $\mathcal{N}_{r}$ the number of linear regions computed by the PWL functions of rectifier network, and $\mathcal{N}_{r}$ is upper and lower bounded by
\begin{align}
\sum_{(j_{1},\ldots,j_{L})\in J,\,(u_{1},\ldots,u_{L})\in U}\prod_{l=1}^{L}R_{lower}\left(j_{l},u_{l}\right)\leq\mathcal{N}_{r}\leq\sum_{(\tilde{j}_{1},\ldots,\tilde{j}_{L})\in\tilde{J}}\prod_{l=1}^{L}R_{upper}\left(\tilde{j}_{l}\right)
\end{align}
where $J=\{(j_{1},\ldots,j_{L})\in\mathbb{Z}^{L}:\ 0\leq j_{l}\leq d_{l}\ \forall l=1,\ldots,L\}$, $U=\{(u_{1},\ldots,u_{L})\in\mathbb{Z}^{L}:\ 0\leq u_{l}\leq j_{l}\ \forall l=1,\ldots,L\}$, $\tilde{J}=\{\tilde{j}_{1},\ldots,\tilde{j}_{L}\in\mathbb{Z}^{L}:\ 0\leq\tilde{j}_{l}\leq\tilde{d}_{l}\ \forall l=1,\ldots,L\}$, and $\tilde{d}_{l}=\min\{n_{0},\tilde{j}_{1},\ldots,\tilde{j}_{l-1},n_{l}\}$. $R_{lower}\left(j_{l},u\right)$ is the number of activation patterns of layer $l$, where $j_{l}$ groups of neurons are active and $u$ groups among them are from the first $r_{rem}^{l}$ groups. $R_{upper}(\tilde{j}_{l})$ is the maximal possible number of linear regions whose corresponding number of active neurons is larger than $\tilde{d}_{l}$ when $\tilde{j}_{l}=\tilde{d}_{l}$ and the least possible number of linear regions with their number of active neurons equal to $\tilde{j}_{l}$ when $0\leq\tilde{j}_{l}<\tilde{d}_{l}$ for $l\in[L]$. The explicit forms of $R_{lower}\left(j_{l},u\right)$ and $R_{upper}\left(\tilde{j}_{l}\right)$ are given in Appendix \ref{E} as well as the recursive relationship between $d_{l}$ and $d_{l+1}$.

\end{theorem}
\begin{IEEEproof}
See Appendix \ref{E}.
\end{IEEEproof} 

This upper bound is more tight on account that the complementary properties between the numbers of active neurons of polyhedrons are considered. The lower bound is improved by counting the activation pattern that some of neurons are non-active. Such lower bound can be applied to more general settings of rectifier networks compared with previous ones \cite{Pascanu2013,Montufar2014,Serra2018}. In particular, upper and lower bounds are equal to  $\left(L+1\right)^{n_{0}}$ when the number of neurons of all the hidden layers is $n_{0}$. The lower bound of Mont\'{u}far et al. is $2^{n_{0}}$ under such setting (See \cite{Montufar2014}, Theorem 4). As the layer of rectifier networks deepens, our result is greatly improved over that of Mont\'{u}far et al.. When $n_{0}=1$, the lower bound in Theorem \ref{theorem3} is tight on any settings of rectifier networks.

\section{Discussion and Conclusions}
The complexity of functions computable by deep neural networks is studies in this paper by counting the number of linear regions of input space. We mainly focused on deep neural networks with piecewise linear activations that is widely used in deep learning and discussed their number of linear regions computed by the PWL functions. 

Firstly, we computed the exact maximal number of linear regions for shallow PLNNs as well as its asymptotic expansion. Furthermore, we derived the upper bound on the number of linear regions for deep PLNNs by bounding the number of subregions partitioned layer by layer. We analyzed the corresponding asymptotic expansions of this upper bounds. It indicated that the structure of neural networks and the non-linearity of activation functions have great influence on the complexity of functions computed by deep PLNNs. The composition of layers results in exponential growth of linear regions compared with shallow counterpart. As layers deeper, the functions computed by deep PLNNs are more expressive. This result provides another perspective on the reason of superior performance of DNNs. 

Moreover, we tightened the upper and lower bounds on the number of linear regions for rectifier networks. The upper bound were lowered down based on the complementary property of the number of active neurons of polyhedrons. The lower bound were improved by considering the activation that some neurons are nonactive.     

In the future works, three aspects are worth studying. Firstly, the bounds provided in this paper still have plenty of room for improvement, especially for rectifier networks due to their special properties. Secondly, how parameter distribution affects the number of linear regions is still unknown as parameters are assumed to be fixed in this paper. Finally, one interesting question is computing the expressiveness of other popular neural network architecture, such as convolutional neural networks.

% or


\appendices
\section{Proof of Theorem \ref{theorem1}}\label{A}
\begin{IEEEproof}
For proof convenience, we begin with a few definitions. For $1\leq d\leq n_{0}$ and $1\leq n\leq n_{1}$, let $\{P_{i,j}\}_{i\in[n],j\in[p]}$ be $n$ groups of $p$ parallel hyperplanes in $d$-dimension space, where $\{P_{i,j}\}$ are mutually parallel if they have the same index $i$ and non-parallel with different $i$. Define $\mathcal{A}_{k}\left(d,n\right)$ as a specific hyperplane arrangement defined by $\mathbb{R}^{d}\backslash\left(\cup_{i\in[k]}P_{i,j} + \cup_{i\in[k+1,n]}P_{i,1}\right)$, $j\in[p],\ k=0,\ldots,n$, where the arrangement has only $k$ groups of $p$ parallel hyperplanes and the remaining $n-k$ hyperplanes have no parallel companions. Let $\mathcal{R}_{k}(d,n)$ be the maximal number of linear regions generated by $\mathcal{A}_{k}\left(d,n\right)$. By definition, $\mathcal{A}_{k}\left(d,n\right)$ is equivalent to $\mathcal{A}$ when $k=n_{1}$, $n=n_{1}$ and $d=n_{0}$. Naturally, the maximal number of linear regions of shallow PLNNs is given by $\mathcal{R}_{n_{1}}\left(n_{0},n_{1}\right)$.
When $p=1$, $\mathcal{R}_{n_{1}}\left(n_{0},n_{1}\right)$ is directly given by $\sum_{n=0}^{n_{0}}\binom{n_{1}}{n}$ \cite{Pascanu2013}, which conforms to Theorem \ref{theorem1}. 
When $p>1$, parallelisms of some hyperplanes of $\mathcal{A}$ need to be taken into account. 

Generally, we can only select at most $n_{1}$ mutually non-parallel hyperplanes from $\mathcal{A}$. Assume that these $n_{1}$ non-parallel hyperplanes are in general positions, which form a hyperplane arrangement $\mathcal{A}_{0}\left(n_{0},n_{1}\right)$ with its maximal number of linear regions equal to $\sum_{n=0}^{n_{1}}\binom{n_{1}}{n}$. If adding a new hyperplane $P$ to $\mathcal{A}_{0}\left(n_{0},n_{1}\right)$, it will intersect with at most $n_{1}-1$ hyperplanes as it must be parallel to one of hyperplanes in $\mathcal{A}_{0}\left(n_{0},n_{1}\right)$. Each intersection is an $(n_{0}-2)$-dimension hyperplane inside $P$. The maximal number of newly partitioned linear regions generated by introducing $P$ is exactly the same with the maximal number of linear regions partitioned by $n_{1}-1$ intersections within $P$ (See \cite{Anthony1999}, Lemma 3.3). The hyperplane arrangement formed by these $n_{1}-1$ intersections is equivalent to $\mathcal{A}_{0}\left(n_{0}-1,n_{1}-1\right)$, and the maximal number of linear regions generated by $\mathcal{A}_{0}\left(n_{0}-1,n_{1}-1\right)$ is given by $\mathcal{R}_{0}\left(n_{0}-1,n_{1}-1\right)$. If adding all the remaining $p-1$ parallel hyperplanes to $\mathcal{A}_{0}\left(n_{0},n_{1}\right)$, newly generated regions are at most $(p-1)\mathcal{R}_{0}\left(n_{0}-1,n_{1}-1\right)$.

Another $n_{1}-1$ groups of parallel hyperplanes in $\mathcal{A}$ can process in a similar way. Adding these groups of parallel hyperplanes one by one, we obtain a recursive relation as follows
\begin{align}\label{recurr}
\mathcal{R}_{k}\left(n_{0},n_{1}\right)=\sum_{n=0}^{n_{0}}\binom{n_{1}}{n}+\sum_{i=0}^{k-1}(p-1)\mathcal{R}_{i}(n_{0}-1, n_{1}-1)
\end{align}
for $k=0,\ldots,n_{1}$. As  $\mathcal{R}\left(\mathcal{A}\right)=\mathcal{R}_{n_{1}}\left(n_{0},n_{1}\right)$, $\mathcal{R}\left(\mathcal{A}\right)$ is recursively derived by replacing $k$ with $n_{1}$ in Eq.\eqref{recurr}. Next, we will show that the explicit form of $\mathcal{R}_{k}\left(n_{0},n_{1}\right)$ is given by
\begin{align}\label{bound}
\mathcal{R}_{k}\left(n_{0},n_{1}\right)=\sum_{i=0}^{k}\binom{k}{i}\left(p-1\right)^{i}\sum_{n=0}^{n_{0}-i}\binom{n_{1}-i}{n}
\end{align}
by induction. 

\emph{Base case} $k=0$. The $n_{0}$-dimension input space is partitioned by $n_{1}$ mutually non-parallel hyperplanes. The maximal number of linear regions is directly given by $\sum_{n=0}^{n_{0}}\binom{n_{1}}{n}$, which is equal to $\mathcal{R}_{0}\left(n_{0},n_{1}\right)$. Hence, the base case holds.

\emph{Induction step}. Assume that $\mathcal{R}_{k-1}\left(n_{0},n_{1}\right)$ satisfies the formula in Eq.\eqref{bound}. Based on the recurrence relation in Eq.\eqref{recurr}, $\mathcal{R}_{k}\left(n_{0},n_{1}\right)$ is written as
\begin{align}
\mathcal{R}_{k}\left(n_{0},n_{1}\right)&=\sum_{n=0}^{n_{0}}\binom{n_{1}}{n}+\sum_{i=0}^{k-1}\sum_{j=0}^{i}\binom{i}{j}\left(p-1\right)^{j+1}\sum_{n=0}^{n_{0}-1-j}\binom{n_{1}-1-j}{n}\nonumber\\
&=\sum_{n=0}^{n_{0}}\binom{n_{1}}{n}+\sum_{i=0}^{k-1}\sum_{j=i}^{k-1}\binom{j}{i}\left(p-1\right)^{i+1}\sum_{n=0}^{n_{0}-1-i}\binom{n_{1}-1-i}{n}.
\end{align}
The coefficient $\sum_{j=i}^{n_{1}-1}\binom{j}{i}$ is given by
\begin{align}
\sum_{j=i}^{k-1}\binom{j}{i}&=\binom{i+1}{i+1}+\binom{i+1}{i}+\binom{i+2}{i}+\cdots+\binom{k-1}{i}\nonumber\\
&=\binom{i+2}{i+1}+\binom{i+2}{i}+\cdots+\binom{k-1}{i}=\binom{k}{i+1}.
\end{align}
Applying the result above, $\mathcal{R}_{k}\left(n_{0},n_{1}\right)$ can be further written as
\begin{align}
\mathcal{R}_{k}\left(n_{0},n_{1}\right)&=\sum_{n=0}^{n_{0}}\binom{n_{1}}{n}+\sum_{i=1}^{k}\binom{k}{i}\left(p-1\right)^{i}\sum_{n=0}^{n_{0}-i}\binom{n_{1}-i}{n}\nonumber\\
&=\sum_{i=0}^{k}\binom{k}{i}\left(p-1\right)^{i}\sum_{n=0}^{n_{0}-i}\binom{n_{1}-i}{n},
\end{align}
which finishes the proof.

Next, we derive asymptotic expansions of $\mathcal{R}\left(\mathcal{A}\right)$ on condition 
that $n_{0}=O(1)$. Substituting $k$ by $n_{1}$, $\mathcal{R}\left(\mathcal{A}\right)$ is given by
\begin{align}\label{bound1}
\mathcal{R}\left(\mathcal{A}\right)=\sum_{i=0}^{n_{1}}\binom{n_{1}}{i}\left(p-1\right)^{i}\sum_{n=0}^{n_{0}-i}\binom{n_{1}-i}{n},
\end{align}

When $n_{1}\leq n_{0}$, we have $\sum_{n=0}^{n_{1}-i}\binom{n_{1}-i}{n}=2^{n_{1}-i}$. Applying this result to Eq.\eqref{bound1}, it follows that $\mathcal{R}\left(\mathcal{A}\right)=\left(p+1\right)^{n_{1}}$. When $n_{1}>n_{0}$, we'll show that $\mathcal{R}\left(\mathcal{A}\right)$ behaves asymptotically as 
$\Theta\left((p+n_{1})^{n_{0}}\right)$. According to Pascanu el al.'s result (See \cite{Pascanu2013}, Proposition 6), it follows that $\sum_{n=0}^{n_{0}}\binom{n_{1}}{n}=\Theta\left(n_{1}^{n_{0}}\right)$. Asymptotically,  $\mathcal{R}\left(\mathcal{A}\right)$ behaves as
\begin{align}
\mathcal{R}\left(\mathcal{A}\right)=\sum_{i=0}^{n_{0}}\binom{n_{1}}{i}\left(p-1\right)^{i}\Theta\left((n_{1}-i)^{n_{0}-i}\right).
\end{align}
Furthermore, $\mathcal{R}\left(\mathcal{A}\right)$ is bounded by
\begin{align}\label{bound2}
\sum_{i=0}^{n_{0}}\binom{n_{1}}{i}\left(p-1\right)^{i}\Theta\left((n_{1}-n_{0})^{n_{0}-i}\right)\leq\mathcal{R}\left(\mathcal{A}\right)\leq\sum_{i=0}^{n_{0}}\binom{n_{1}}{i}\left(p-1\right)^{i}\Theta\left((n_{1})^{n_{0}-i}\right).
\end{align}
The upper and lower bounds in Eq.\eqref{bound2} can be further written as
\begin{align}
\Theta\left((p-1+n_{1}-n_{0})^{n_{0}}\right)\leq\mathcal{R}\left(\mathcal{A}\right)\leq\Theta\left((p-1+n_{1})^{n_{0}}\right).
\end{align}
Since $n_{0}=O(1)$, it follows that $\mathcal{R}\left(\mathcal{A}\right)=\Theta\left((p+n_{1})^{n_{0}}\right)$.
\end{IEEEproof}

\section{Proof of Theorem \ref{theorem2}}\label{B}
\begin{IEEEproof}
For a single layer, the maximal number of linear regions is exactly given by $\mathcal{R}\left(\mathcal{A}\right)$, which is equal to $\mathcal{R}\left(n_{0},n_{1}\right)$. For $l>1$, suppose that $R$ is a linear region of layer $l-1$ and $R\in\mathcal{R}^{l-1}$ corresponds to the activation vector $\mathbf{s}^{l-1}\in\mathcal{S}^{l-1}$. It would be partitioned by a hyperplane set $P_{\mathbf{s}_{l-1}}$ intersected with $R$. The maximal number of linear regions partitioned by these intersected hyperplanes is upper bounded by the result of Theorem \ref{theorem1}. Moreover, the degree of freedom of these hyperplanes is constrained by the rank of $\hat{\mathbf{W}}^{l}$ (See \cite{Serra2018}, Lemma 4). Let $d_{l}$ be the maximal rank of $\hat{\mathbf{W}}^{l}$ and $d_{l}=\min\{n_{0},n_{1},\ldots,n_{l}\}$. Hence, $R$ has at most $\mathcal{R}\left(d_{l}, n_{l}\right)$ subregions. The recursive relationship between different layers is written as
\begin{align}\label{upper1}
\mathcal{N}^{l}\ \left\{\begin{array}{cc}
\leq\mathcal{R}\left(d_{l},n_{l}\right)\times\mathcal{N}^{l-1}, & \mathrm{if}\ 2\leq l\leq L,\\
=\mathcal{R}\left(d_{1},n_{1}\right), & \mathrm{if}\ l=1.
\end{array}\right.
\end{align}
By unpacking the Eq.\eqref{upper1}, the upper bound on the number of linear regions up to layer $L$ is given by
\begin{align}\label{bound3}
\mathcal{N}^{L}\leq\prod_{l=1}^{L}\mathcal{R}\left(d_{l},n_{l}\right).
\end{align}
\end{IEEEproof}

\section{Analysis of Theorem \ref{theorem2}}\label{C}
The asymptotic expansions of upper bound in Theorem \ref{theorem2} are derived to illustrate key factors that affect the number of linear regions computable by deep models. 

\begin{lemma}\label{lemma2}
For $n_{l}\geq d_{l}\geq1$, $\mathcal{R}\left(d_{l},n_{l}\right)$ is upper bounded by
\begin{align}
\mathcal{R}\left(d_{l},n_{l}\right)\leq\left[p-1+e\left(n_{l}-d_{l}+1\right)\right]^{d_{l}}.
\end{align}
\end{lemma}
\begin{IEEEproof}
According to Anthony el al.'s theorem (See \cite{Anthony1999}, Theorem 3.7), we have 
\begin{align}
\sum_{i=0}^{d}\binom{m}{i}<\left(\frac{em}{d}\right)^{d}
\end{align}
for $m\geq d\geq 1$. Applying this theorem, $\mathcal{R}\left(d_{l},n_{l}\right)$ can be rewritten as
\begin{align}
\mathcal{R}\left(d_{l},n_{l}\right)&\leq\sum_{i=0}^{d_{l}-1}\binom{d_{l}}{i}\left(p-1\right)^{i}\left[\frac{e(n_{l}-i)}{d_{l}-i}\right]^{d_{l}-i}+(p-1)^{d_{l}}\nonumber\\
&\leq\sum_{i=0}^{d_{l}-1}\binom{d_{l}}{i}\left(p-1\right)^{i}\left[e(n_{l}-d_{l}+1)\right]^{d_{l}-i}+(p-1)^{d_{l}}\nonumber\\
&=\left[p-1+e\left(n_{l}-d_{l}+1\right)\right]^{d_{l}}
\end{align}
for $n_{l}\geq d_{l}\geq 1$. 
\end{IEEEproof}

Substituting this bound into Eq.\eqref{bound3}, $\mathcal{N}^{L}$ is upper bounded by
\begin{align}
\mathcal{N}^{L}&\leq\prod_{l=1}^{L}\left[p-1+e\left(n_{l}-d_{l}+1\right)\right]^{d_{l}}\leq\left[p-1+\sum_{l=1}^{L}e\left(n_{l}-d_{l}+1\right)/L\right]^{\sum_{l=1}^{L}d_{l}}.
\end{align}
Define the effective length as $\bar{L}=\frac{1}{n_{0}}\sum_{l=1}^{L}d_{i}$. It follows that the upper bound of $\mathcal{N}^{L}$ is rewritten as
\begin{align}\label{bound4}
\mathcal{N}^{L}\leq\left[p-1+e\left(N/L-n_{0}\bar{L}/L+1\right)\right]^{\bar{L}n_{0}}.
\end{align}
Consider a shallow PLNN with $p$ pieces of linear activations, $N$ hidden neurons and $n_{0}$-dimension input. According to Lemma \ref{lemma2}, its number of linear regions is upper bounded by
\begin{align}
\mathcal{N}^{1}\leq\left[p-1+e\left(N-n_{0}+1\right)\right]^{n_{0}}.
\end{align}
Hence, the asymptotic expansion of upper bound for shallow PLNNs is  $\mathcal{N}^{1}=O\left(p+eN\right)^{n_{0}}$, which is similar to the asymptotic expansion given by Theorem \ref{theorem1} except some different parameters. Furthermore, the effective length $\bar{L}$ behaves asymptotically as $\bar{n}/n_{0}\Theta\left(L\right)$, where $1\leq\bar{n}\leq n_{0}$ is a scaling factor. It follows that the asymptotic expansion of upper for deep PLNNs is $\mathcal{N}^{L}=O\left(p+eN/L\right)^{\bar{n}L}$.
For both of deep and shallow models, the non-linearity introduced by hidden units has also greatly affected the expressive power along with the number of hidden neurons. While, the upper bound on the number of linear regions for deep models grows exponentially over that of shallow models as the layer $L$ increases. It can generate much more linear regions than their shallow counterparts due to recursive partition of the linear regions of previous layer. 

\section{Proof of Lemma \ref{lemma1}}\label{D}
\begin{IEEEproof}
Firstly, we prove that each polyhedron has a counterpart with their active neurons added up to $m$. Given an input $\mathbf{x}$ in general position, the activation pattern is determined by a set of linear inequalities that certain hidden neurons are active if $\mathbf{w}_{i}^{T}\mathbf{x}+b_{i}>0$ or nonactive otherwise. Next, we show that the inequality $\mathbf{w}_{i}^{T}\mathbf{x}+b_{i}>0$ determining whether the neuron is active is equivalent to $\mathbf{w}_{i}^{T}\mathbf{x}>0$. Suppose that $\mathbf{x}$ belongs to one of polyhedrons and the first $m_{1}$ neurons are active, i.e., $\mathbf{w}_{i}^{T}\mathbf{x}+b_{i}>0$ for $i\in[m_{1}]$. For $\alpha\in\mathbb{R}^{+}$, $\alpha\mathbf{x}$ also belongs to the polyhedron, and we have $\alpha\mathbf{w}_{i}^{T}\mathbf{x}+b_{i}>0$ as well. Therefore, the value of $b_{i}$ doesn't affect the sign of such linear inequality, and these $m_{1}$ inequalities are equivalent to $\mathbf{w}_{i}^{T}\mathbf{x}>0$. If we flip the sign of every element of $\mathbf{x}$, $m_{1}$ linear inequalities is converted to $\mathbf{w}_{i}^{T}\mathbf{x}<0$. The remaining $m-m_{1}$ linear inequalities are larger than zero in turn. These sign-flipped linear inequalities correspond to a polyhedron that $m-m_{1}$ hidden neurons are active. As a result, by flipping the sign of every element of $\mathbf{x}$, every polyhedron has a counterpart with their numbers of active hidden neurons added up to $m$.

Since biases have no effects on the number of active neurons in the case of polyhedrons, the maximal number of polyhedrons is equivalent to the maximal number of cones partitioned by a set of $m$ hyperplanes $\{\mathbf{w}_{i}^T\mathbf{x}=0\}_{i\in[m]}$. If $\{\mathbf{w}_{i}\}_{i\in[m]}$ are linearly independent, the maximal number of cones is given by $2\sum_{n=0}^{n_{0}-1}\binom{m-1}{n}$ according to Anthony's result (See \cite{Anthony1999}, Lemma 3.3). More generally, if $r$ is the rank of $\mathbf{W}$, a subset of $\{\mathbf{w}_{i}\}_{i\in[m]}$ has at most $r$ linearly independent vectors, and such result is rewritten as $2\sum_{n=0}^{r-1}\binom{m-1}{n}$ (See \cite{Serra2018}, Lemma 4).   
\end{IEEEproof}

\section{Proof of Theorem \ref{theorem3}}\label{E}

Firstly, we derive the explicit form of $R_{upper}\left(\tilde{j}_{l}\right)$. According to Serra et al.'s theorem (See \cite{Serra2018}, Theorem 1), the effective dimension of input space up to layer $l$ is given by $\tilde{d}_{l}=\min\{\tilde{d}_{l-1},\ldots,n_{l}\}$. If the number of active neurons of layer $l$ is larger than $\tilde{d}_{l}$, we have $\tilde{d}_{l+1}=\tilde{d}_{l}$. If the number of active neurons is lower than $\tilde{d}_{l}$, $\tilde{d}_{l+1}$ must be lower than $\tilde{d}_{l}$. To upper bound the number of linear regions of rectifier networks, we need to activate as many neurons as possible by taking the complementary constrain of Lemma \ref{lemma1} into account. 

If $n_{l}=\tilde{d}_{l}$, all the linear regions are polyhedrons if $\mathcal{N}_{r}$ is maximized, and $R_{upper}(\tilde{j}_{l})=\binom{n_{l}}{\tilde{j}_{l}}$. The effective dimension of input space of next layer is $\tilde{d}_{l+1}=\tilde{j}_{l}$. When $\tilde{d}_{l}<n_{l}\leq 2\tilde{d}_{l}$, the least number of active neurons is $n_{l}-\tilde{d}_{l}$ according to Serra et al.'s theorem (See \cite{Serra2018}, Theorem 1). The maximal number of linear regions with $n_{l}-\tilde{d}_{l}$ active neurons is $\binom{n_{l}}{n_{l}-\tilde{d}_{l}}$. While, such result is possible when the following inequality is satisfied
\begin{align}\label{inequality}
\sum_{n=n_{l}-\tilde{d}_{l}}^{\tilde{d}_{l}}\binom{n_{l}}{n}\geq2\sum_{n=0}^{\tilde{d}_{l}-1}\binom{n_{l}-1}{n}.
\end{align}  
This inequality indicates that it is possible the number of active neurons all the polyhedrons correspond to is not lower than $n_{l}-\tilde{d}_{l}$. If such inequality isn't satisfied, we must have a polyhedron that its number of active neurons is larger than $\tilde{d}_{l}$ or lower than $n_{l}-\tilde{d}_{l}$. If the polyhedron has more than $\tilde{d}_{l}$ active neurons, it must have a complementary polyhedron that corresponding number of active neurons is lower than $n_{l}-\tilde{d}_{l}$. Either way, we must have a polyhedron that its number of active neurons is lower than $n_{l}-\tilde{d}_{l}$. To upper bound the number of linear regions of rectifier networks, we should make the number of such polyhedrons least. Therefore, the linear regions with the number of active neurons between $n_{l}-\tilde{d}_{l}$ and $\tilde{d}_{l}$ should accommodate as many polyhedrons as possible. Denote by $\Delta R_{1,\tilde{d}_{l}}=2\sum_{n=0}^{\tilde{d}_{l}-1}\binom{n_{l}-1}{n}-\sum_{n=n_{l}-\tilde{d}_{l}}^{\tilde{d}_{l}}\binom{n_{l}}{n}$ the difference between the capacity and the number of polyhedrons. If $\Delta R_{1,\tilde{d}_{l}}>0$, the number of polyhedrons exceeds the capacity, and there must be a polyhedron with its number of active neurons beyond $\tilde{d}_{l}$ or lower than $n_{l}-\tilde{d}_{l}$. According to complementary property of Lemma \ref{lemma1}, the least number of polyhedrons with their number of active neurons lower than $n_{l}-\tilde{d}_{l}$ is $\Delta R_{1,\tilde{d}_{l}}/2$. Therefore, the maximal number of linear regions with their number of active neurons larger than $n_{l}-\tilde{d}_{l}$ is $\sum_{n=n_{l}-\tilde{d}_{l}}^{n_{l}}\binom{n_{l}}{n}-\Delta R_{1,\tilde{d}_{l}}/2$. 

Base on the analysis above, $R_{upper}(\tilde{j}_{l})$ is given by $\sum_{n=\tilde{d}_{l}}^{n_{l}}\binom{n_{l}}{n}-f_{relu}(\Delta R_{1,\tilde{d}_{l}}/2)$ when $\tilde{j}_{l}=\tilde{d}_{l}$ and $\binom{n_{l}}{\tilde{j}_{l}}$ when $n_{l}-\tilde{d}_{l}\leq j_{l}<\tilde{d}_{l}$. Then, the number of remaining polyhedrons is $\Delta R_{1,\tilde{d}_{l}}/2$ when $\tilde{j}_{l}<n_{l}-\tilde{d}_{l}$. Let $c_{1}(\tilde{j}_{l})=\Delta R_{1,\tilde{d}_{l}}/2-\sum_{n=\tilde{j}_{l}}^{n_{l}-\tilde{d}_{l}-1}\binom{n_{l}}{n}$ be the number of remaining ls by subtracting the maximal number of polyhedrons with their active neurons ranging from $\tilde{j}_{l}$ to $n_{l}-\tilde{j}_{l}$. If $c_{1}(\tilde{j}_{l})\geq 0$, the maximal number of polyhedrons with their active neurons equal to $\tilde{j}_{l}$ is achievable, and we have  $R_{upper}(\tilde{j}_{l})=\binom{n_{l}}{\tilde{j}_{l}}$. If $c_{1}(\tilde{j}_{l})<0$ and $c_{1}(\tilde{j}_{l}+1)>0$, it's not possible to maximize the number of linear regions their active neurons equal to $\tilde{j}_{l}$, and $R_{upper}(\tilde{j}_{l})=c_{1}(\tilde{j}_{l}+1)$. If $c_{1}(\tilde{j}_{l}+1)\leq 0$, we have no remaining polyhedrons and $R_{upper}(\tilde{j}_{l})=0$. Aggregating these results, the explicit form of $R_{upper}(\tilde{j}_{l})$ is given by
\begin{align}\label{upper}
R_{upper}\left(\tilde{j}_{l}\right)=\left\{\begin{array}{cc}
\sum_{n=\tilde{d}_{l}}^{n_{l}}\binom{n_{l}}{n}-f_{relu}\left(\Delta R_{1,\tilde{d}_{l}}/2\right) &\mathrm{if}\ \tilde{j}_{l}=\tilde{d}_{l}\\
\binom{n_{l}}{\tilde{j}_{l}}&\mathrm{if}\ n_{l}-\tilde{d}_{l}\leq\tilde{j}_{l}< \tilde{d}_{l}\ |\left[\ 0<\tilde{j}_{l}<n_{l}-\tilde{d}_{l}\ \&\ c_{1}\left(\tilde{j}_{l}\right)\geq 0\right] \\
c_{1}\left(\tilde{j}_{l}+1\right)&\mathrm{if}\ 0<\tilde{j}_{l}<n_{l}-\tilde{d}_{l}\ \&\ c_{1}\left(\tilde{j}_{l}\right)< 0\ \&\ c_{1}\left(\tilde{j}_{l}+1\right)>0\\
0 &\mathrm{if}\ c_{1}\left(\tilde{j}_{l}+1\right)\leq 0
\end{array}\right.
\end{align}
When $n_{l}>2\tilde{d}_{l}$, the analysis is similar to the case when $\tilde{d}_{l}<n_{l}\leq2\tilde{d}_{l}$ with minor modifications. The inequality in Eq.\eqref{inequality} is reformulated as
\begin{align}
\sum_{n=\tilde{d}_{l}}^{n_{l}-\tilde{d}_{l}}\binom{n_{l}}{n}\geq2\sum_{n=0}^{\tilde{d}_{l}-1}\binom{n_{l}-1}{n}.
\end{align}  
As a result, the difference is rewritten as $\Delta R_{2,\tilde{d}_{l}}=2\sum_{n=0}^{\tilde{d}_{l}-1}\binom{n_{l}-1}{n}-\sum_{n=\tilde{d}_{l}}^{n_{l}-\tilde{d}_{l}}\binom{n_{l}}{n}$. In a similar way, $R_{upper}(\tilde{j}_{l})$ is given by
\begin{align}
R_{upper}\left(\tilde{j}_{l}\right)=\left\{\begin{array}{cc}
\sum_{n=n_{l}-\tilde{d}_{l}}^{n_{l}}\binom{n_{l}}{n}-f_{relu}\left(\Delta R_{2,\tilde{d}_{l}}/2\right) &\mathrm{if}\ \tilde{j}_{l}=\tilde{d}_{l}\\
\binom{n_{l}}{\tilde{j}_{l}}&\mathrm{if}\ \ 0<\tilde{j}_{l}<\tilde{d}_{l}\ \&\ c_{2}\left(\tilde{j}_{l}\right)\geq 0 \\
c_{2}\left(\tilde{j}_{l}+1\right)&\mathrm{if}\ 0<\tilde{j}_{l}<\tilde{d}_{l}\ \&\ c_{2}\left(\tilde{j}_{l}\right)< 0\ \&\ c_{2}\left(\tilde{j}_{l}+1\right)>0\\
0 &\mathrm{if}\ c_{2}\left(\tilde{j}_{l}+1\right)\leq 0
\end{array}\right.
\end{align}
where $c_{2}\left(\tilde{j}_{l}\right)=\Delta R_{2,\tilde{d}_{l}}/2-\sum_{n=n_{l}-\tilde{d}_{l}+1}^{n_{l}-\tilde{j}_{l}}\binom{n_{l}}{n}$. Given a set $\{\tilde{j}_{1},\ldots,\tilde{j}_{L}\}$, $\prod_{l=1}^{L}R_{upper}\left(\tilde{j}_{l}\right)$ is corresponding maximal number of linear regions under such configuration. The effective dimension satisfies $\tilde{d}_{l}=\min\{n_{0},\tilde{j}_{1},\ldots,\tilde{j}_{l-1},n_{l}\}$. Adding up all the possible configurations of $(\tilde{j}_{1},\ldots,\tilde{j}_{L})$, we derive the upper bound on the number of linear regions of rectifier networks, which is given by
\begin{align}
\sum_{(\tilde{j}_{1},\ldots,\tilde{j}_{L})\in\tilde{J}}\prod_{l=1}^{L}R_{upper}\left(\tilde{j}_{l}\right)
\end{align}
where $\tilde{J}=\{(\tilde{j}_{1},\ldots,\tilde{j}_{L})\in\mathbb{Z}^{L}:\ 0\leq\tilde{j}_{l}\leq\tilde{d}_{l}\ \forall l=1,\ldots,L\}$. 

Next, we propose a modified construction strategy to improve the lower bound on the number of linear regions of rectifier networks. According to Mont\'{u}far et al.'s theorem (See \cite{Montufar2014}, theorem 4), every hidden layer connects to a fictitious intermediary layer as inputs, and their numbers of units are $\tilde{d}$. The neurons of hidden layers up to $L-1$ layer is partitioned into $\tilde{d}$ independent groups, and each group is connected to a separate intermediary unit. We can construct a zigzag pattern to map all the inputs from intermediary units to single output range. The number of activation patterns is equal to the number of slopes. Next, we give a detailed analysis on the construction and the neurons of last hidden layer aren't considered. 

If each group has more than two neurons, we can construct a zigzag pattern that at least one neuron is active, and the maximal number of activation patterns is equal to the number of neurons in each group plus one (See \cite{Serra2018}, Theorem 7). The dimension of next intermediary layer isn't changed as the outputs of all the groups are nonzero. If the number of neurons of each group is lower than or equal to two, the constructed zigzag pattern maximizing the number of activation patterns must have an activation pattern that all the neurons are nonactive for each group. The dimension of next intermediary layer is decreased if all the neurons of some groups are nonactive. Therefore, $d_{l}$ is dependent on the number of active groups of previous hidden layer. Based on principle, the explicit form of $R_{lower}\left(j_{l},u\right)$ and the recursive relationship between $d_{l}$ and $d_{l+1}$ are given in the next. 
 
When $n_{l}=\tilde{d}$, each group of hidden layer $l$ has only one neuron. The dimension of connected intermediary layer is assumed to be $d_{l}$. Furthermore, $d_{l}$ is number of nonzero intermediary units. Therefore, only $d_{l}$ groups would be active, and these active groups are from the first $d_{l}\leq\tilde{d}$ groups. When $n_{l}=\tilde{d}$, all the neurons are evenly partitioned, i.e., $r_{rem}^{l}=0$. When $u=0$, $R_{lower}\left(\tilde{j}_{l}, u\right)$ is simply given by $\binom{d_{l}}{j_{l}}$. The dimension of next intermediary layer is determined by the number of active groups, which is given by $d_{l+1}=j_{l}$. When $u>0$, we have $R_{lower}\left(j_{l}, u\right)=0$ as $r_{rem}^{l}=0$.

When $n_{l}\geq3\tilde{d}$, each group has more than two neurons. The number of active groups is $d_{l}$. For each active group, we can construct a zigzag pattern with at least one neuron active. The first $r_{rem}^{l}$ groups have $\lfloor n_{l}/\tilde{d}\rfloor+2$ activation patterns, and the rest have $\lfloor n_{l}/\tilde{d}\rfloor+1$ activation patterns. If $d_{l}\leq r_{rem}^{l}$, all the active groups are from the first $r_{rem}^{l}$ groups. When $u=j_{l}$, $R_{lower}\left(j_{l},u\right)$ is given by 
\begin{align}\label{lower}
R_{lower}\left(j_{l},u\right)=\binom{d_{l}}{j_{l}}\left(\lfloor n_{l}/\tilde{d}\rfloor+1\right)^{j_{l}}, 
\end{align}
For $0\leq u<j_{l}$, $R_{lower}\left(j_{l},u\right)=0$. If $d_{l}>r_{rem}^{l}$, the next $d_{l}-r_{rem}^{l}$ groups would be active as well. In this case, $R_{lower}\left(j_{l},u\right)$ is rewritten as
\begin{align}\label{lower1}
R_{lower}\left(j_{l},u\right)=
\binom{r_{rem}^{l}}{u}\left(\lfloor n_{l}/\tilde{d}\rfloor+1\right)^{u}\binom{d_{l}-r_{rem}^{l}}{j_{l}-u}\left(\lfloor n_{l}/\tilde{d}\rfloor\right)^{j_{l}-u}.
\end{align}      
Since every group has at least one neuron active, the dimension of next intermediary layer is unchanged, i.e., $d_{l+1}=d_{l}$. 

When $2<n_{l}/\tilde{d}< 3$, only the first $r_{rem}^{l}$ groups have three neurons and the rest have two neurons. The zigzag patterns constructed in the groups of two neurons would not guarantee all the neurons active. The dimension of next intermediary layer may be decreased. If $d_{l}\leq r_{rem}^{l}$, all the active groups from the first $r_{rem}^{l}$ groups and they have at least one neuron active. When $u=j_{l}$, $R_{lower}\left(\tilde{j}_{l},u\right)$ is given by Eq.\eqref{lower}. When $0\leq u<j_{l}$, $R_{lower}\left(j_{l},u\right)=0$. The dimension of next intermediary layer is unchanged, i.e., $d_{l+1}=d_{l}$. If $d_{l}>r_{rem}^{l}$, some groups with two neurons would be active, and $R_{lower}\left(j_{l},u\right)$ is the same with Eq.\eqref{lower1}. Some intermediary units of next layer may be zero as the groups with two neurons have the activation pattern that all the neurons are nonactive. The dimension $d_{l+1}$ is determined by the number of active groups with two neurons, which is equal to $d_{l+1}=r_{rem}^{l}+\min\{j_{l}-u,d_{l}-r_{rem}^{l}\}$ for $u\in[j_{l}]$. 

When $1<n_{l}/\tilde{d}\leq 2$, every group has one or two neurons. The activation patterns induced by the zigzag pattern constructed in these groups must have the mode that all the neurons are nonactive. Similarly, if $d_{l}\leq r_{rem}^{l}$, all the active groups have two neurons and $R_{lower}\left(j_{l},u\right)=\binom{d_{l}}{j_{l}}2^{j_{l}}$ when $u=j_{l}$. For $0\leq u<j_{l}$, $R_{lower}\left(j_{l},u\right)=0$. If $d_{l}>r_{rem}^{l}$, some active groups may have just one neurons, and $R_{lower}\left(j_{l}\right)$ is given by
\begin{align}
R_{lower}\left(j_{l},u\right)=
\binom{r_{rem}^{l}}{u}\binom{d_{l}-r_{rem}^{l}}{j_{l}-u}2^{u}.
\end{align}
The dimension of next intermediary is also equal to the number of active groups, i.e., $d_{l+1}=j_{l}$.

The neurons in the last layer aren't partitioned and $r_{rem}^{l}=0$. These neurons act on a $d_{L}$-dimension cube. The number of activation patterns generated is equivalent to the partition in the whole $d_{L}$-dimension space.  According to Zaslavsky' theorem \cite{Zaslavsky1975}, $R_{lower}\left(j_{L},u\right)$ is given by $\binom{n_{L}}{j_{L}}$ when $u=0$. For $0<u\leq j_{L}$, $R_{lower}\left(j_{L},u\right)$ is simply zero. 

\section*{Acknowledgment}

The authors would like to thank...

\ifCLASSOPTIONcaptionsoff
  \newpage
\fi

\end{document}